\documentclass[11pt]{article}

\usepackage[preprint]{acl}

\usepackage{times}
\usepackage{latexsym}

\usepackage[T1]{fontenc}

\usepackage[utf8]{inputenc}

\usepackage{microtype}

\usepackage{inconsolata}

\usepackage{graphicx}

\usepackage{multirow}
\usepackage{stmaryrd}
\usepackage{xurl}

\title{TEGRA: Text Encoding With Graph and Retrieval Augmentation for Misinformation Detection}

\author{
 \textbf{G\'eraud Faye\textsuperscript{1,2}},
 \textbf{Wassila Ouerdane\textsuperscript{2}},
 \textbf{Guillaume Gadek\textsuperscript{1}},
 \\
 \textbf{Sylvain Gatepaille\textsuperscript{1}},
 \textbf{C\'eline Hudelot\textsuperscript{2}}
\\
\\
 \textsuperscript{1}Airbus Defence and Space,
 \textsuperscript{2}Universit\'e Paris-Saclay, CentraleSupélec, MICS,
}

\begin{document}
\maketitle
\begin{abstract}
Misinformation detection is a critical task that can benefit significantly from the integration of external knowledge, much like manual fact-checking. In this work, we propose a novel method for representing textual documents that facilitates the incorporation of information from a knowledge base. Our approach, Text Encoding with Graph (TEG), processes documents by extracting structured information in the form of a graph and encoding both the text and the graph for classification purposes. Through extensive experiments, we demonstrate that this hybrid representation enhances misinformation detection performance compared to using language models alone. Furthermore, we introduce TEGRA, an extension of our framework that integrates domain-specific knowledge, further enhancing classification accuracy in most cases.
\end{abstract}

\section{Introduction}
\label{sec:intro}

Automatic misinformation detection is a great challenge in the current digital era, as an increasing number of people rely on online platforms for news consumption\footnote{\url{https://reutersinstitute.politics.ox.ac.uk/digital-news-report/2024}}. Other approaches, such as media literacy, are equally important but demand greater effort to be effective against all types of misinformation. Misinformation can take several forms~\citep{typology}, ranging from fake news to clickbait. The most impactful types of misinformation are fake news and rumors, which significantly impact the reputation of public figures, influence the political climate, and even cause financial losses for individuals and companies.

To detect misinformation, humans rely on personal or external knowledge to verify whether a piece of news is true or false. This process requires an understanding of the text and the ability to link facts with external sources backing them. Automatic misinformation detection systems, on the other hand,  typically rely on supervised machine learning models, which struggle to structure and utilise explicit knowledge. Instead, they tend to store the information implicitly within the weights of a neural network~\cite{petroni-etal-2019-language}. A recent general challenge~\citep{10721277} is to hybridize external symbolic knowledge with neural models to achieve better performance and improved knowledge traceability on various knowledge-intensive tasks. Incorporating external knowledge, whether structured or unstructured, appears essential for effective misinformation detection, yet it remains a significant challenge in a general setting.

In this paper, we propose a new methodology for embedding texts by incorporating additional factual features into the classification process, while simultaneously extracting information from the text, which enables the future valorization of the extracted knowledge from the texts. Our main contributions are the following. We propose Text Encoding with Graph (TEG), an approach that combines semantic and factual embeddings by leveraging both the text and an OpenIE graph representation, which captures entities, properties, and actions from the text. This method improves performance in fake news detection and could also benefit other knowledge-based tasks. We demonstrate that the granularity of OpenIE graphs enables the easy addition of information related to extracted entities and properties. By enriching the graph representation with external knowledge in the TEG with Retrieval Augmentation (TEGRA) framework, we achieve additional performance improvements.

The paper is structured as follows. Section~\ref{sec:literature} provides an overview of the related literature, followed by a description of the TEG and TEGRA approaches in Section~\ref{sec:methodo}. The evaluation process and results are discussed in Sections~\ref{sec:eval} and~\ref{sec:results}, respectively, before we conclude in Section~\ref{sec:conclusion}.


\section{Related work}
\label{sec:literature}


\paragraph{General Misinformation Detection.}

Research on misinformation detection broadly utilizes two approaches: \textit{fact-checking} and \textit{text classification}. Fact-checking, often applied at the claim level (short, factual assertions), focuses on assessing the veracity of information, while text classification evaluates entire documents holistically, taking into account both factual and stylistic characteristics.

Recent approaches increasingly leverage Large Language Models (LLMs) and retrieval-augmented generation (RAG) techniques, demonstrating superior performance compared to earlier fine-tuning methods~\citep{10317251}. These techniques have proven effective in various domains, including medical~\citep{kotonya-toni-2020-explainable-automated} and scientific claims~\citep{wadden-etal-2020-fact}, and are closely related to hallucination detection in generated content~\citep{Wang2023FactcheckGPTEF,li-etal-2024-self}. Early text classification efforts relied on simpler models such as N-grams, LSTMs, and TF-IDF. However, the advent of transformer-based models, particularly BERT~\citep{devlin_bert_2019}, established new performance baselines, prompting research focused on enhancing these models for real-world deployment and robustness. \citep{pelrine_2021_surprising}.

\paragraph{Knowledge Incorporation for Misinformation Detection.}

A growing body of work explores the incorporation of external knowledge to enhance misinformation detection, mimicking the human verification process. Early approaches, such as ERNIE~\citep{zhang-etal-2019-ernie} and K-BERT~\citep{kbert}, paved the way for more recent strategies that leverage LLMs, including RAG and Cache-Augmented Generation (CAG)~\citep{chan2024dontragcacheaugmentedgeneration}. However, these approaches often struggle with the challenge of grounding generated or classified content in verifiable, external sources, a crucial requirement for industrial applications that demand high reliability. In addition to these general approaches, recent studies have incorporated knowledge graphs (KGs) to represent relationships between entities and concepts explicitly.

Several strategies exist for integrating KG information. Some approaches directly augment the input text by incorporating knowledge graph embeddings or Wikipedia descriptions, as exemplified by CompareNet~\citep{hu-etal-2021-compare} and DDGCN~\citep{sun_ddgcn_2022}. Others focus on guiding embedding layers, such as in DETERRENT~\citep{cui_deterrent_2020}, where KG entity embeddings are incorporated into the attention mechanism. Open Information Extraction (OpenIE) provides a means to transform text into graph structures resembling KGs, allowing for structured knowledge integration. DTN~\citep{liu_2021_dtn} and Karlapalem et al.~\citep{karlapalem_incorporating_2021} exemplify this approach, which combines textual and graph embeddings to improve performance. Furthermore, architectures such as DeClarE~\citep{popat_declare_2018} and Kaplan et al.~\citep{10286288} utilize related documents and graph embeddings to assess credibility and detect misinformation, respectively. Recent entity-centric graph approaches, such as KAPALM~\citep{ma-etal-2023-kapalm}, build upon these principles. While these KG-enhanced methods offer improvements, they often lack the formal reasoning capabilities needed for rigorous verification and integration into automated systems.


\section{Proposed approach: Text Encoding with Graph and Retrieval Augmentation}
\label{sec:methodo}

Previous approaches that use graph structures to represent text are either at a granularity that does not allow for the addition of information about entities or fail to fully translate the unstructured knowledge in the text into a structured format. Contrary to approaches for misinformation detection presented above, we explicitly convert press articles into OpenIE graphs to generate more fact-centered embeddings, using only the text as input without relying on additional metadata. This representation allows for the incorporation of more factual features in text classification by incorporating features inspired by fact-checking (which is not performed in the text classification task). Our method of incorporating knowledge from other documents is more granular, adding information specifically related to identified entities, rather than incorporating details from entire documents or topics. Because of this, our approach can be easily applied to any knowledge-intensive task that requires enriched text embeddings, not just for misinformation detection. In addition, the proposed approach enables the leveraging of knowledge contained in fake news articles, which is not available or utilized in previous knowledge-based approaches, to enhance their detection at inference.

\begin{figure*}[ht!]
    \centering
    \includegraphics[width=\linewidth]{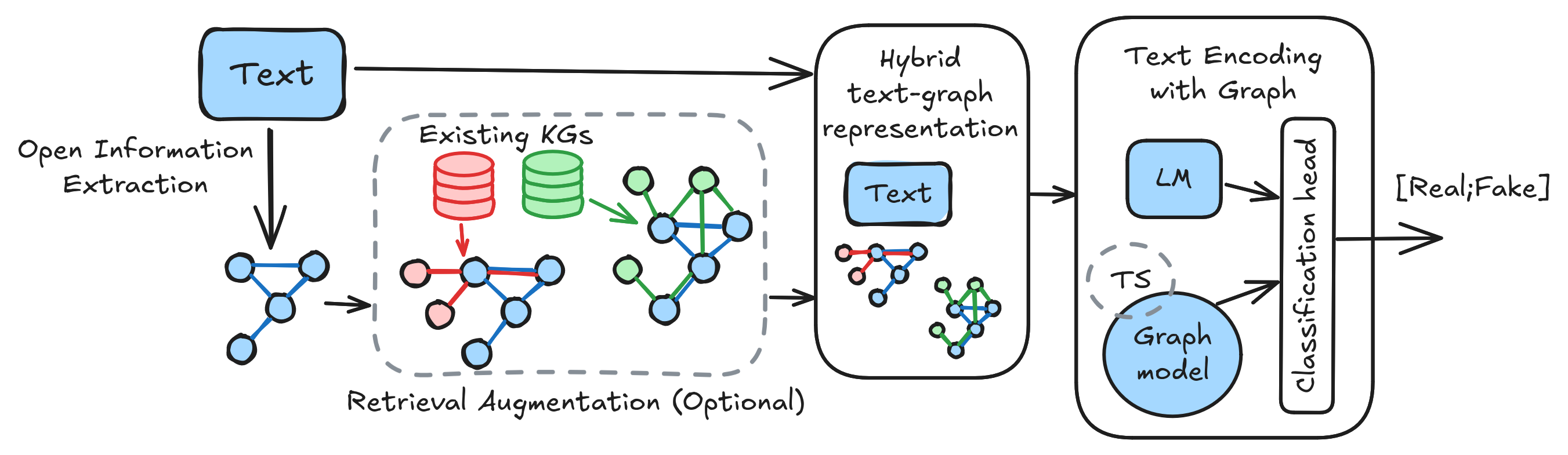}
    \caption{The global processing architecture of TEG. Each component can be swapped with a similar component with the same function.}
    \label{fig:global_archi}
\end{figure*}

Our main idea in the Text Encoding with Graph (TEG) method is to process both the text and an OpenIE graph extracted from the text for classification (see Figure~\ref{fig:global_archi} for the overall processing pipeline, with detailed steps outlined in this section). Due to its structure, this graph-based representation can be more easily linked to knowledge expressed as triples, enabling the integration of external contextual information. Firstly, we will explain what motivated the use of OpenIE graphs in Subsection~\ref{sec:motivation}. Then, in Subsection~\ref{sec:data_processing}, the construction of the hybrid text-graph representation is explained, some methods to incorporate external knowledge are given in Subsection~\ref{sec:knowledge_incoprporation}, and the proposed neural models are detailed in Subsection~\ref{sec:model_processing}.

\subsection{Motivation of using OpenIE graphs for structured representation of data}
\label{sec:motivation}

Structuring data in a human-understandable way has been a long-standing challenge~\cite {9088989}. Structured data extraction achieves high performance and ensures that the extracted information adheres to a predefined ontology, making it compatible with an existing knowledge graph. However, these graphs can be limited, as all types of triples must be defined at an early stage when constructing the base, and may only be modified with great effort later.

In contrast, Open Information Extraction (OpenIE) structures the text globally by extracting all entities and relations in a broader way to represent all actions and properties depicted in the text.

The primary motivation for using OpenIE graphs is their ability to exhaustively represent the information in a text in a structured format. Their clarity makes them an ideal choice for organizing textual content. Additionally, their predominantly entity-centric nature facilitates integration with other entity-focused graph types.

Other graph-based approaches for structuring text, such as K-graphs, syntactic trees, or sentence-based graphs, tend to have structures that differ significantly from knowledge graphs, making it challenging to integrate information from multiple sources or external knowledge bases.

\subsection{Constructing the hybrid text-graph representation of the textual document}
\label{sec:data_processing}

Texts are processed with an OpenIE model to construct a graphical representation of the text. The task was introduced by~\citep{10.1145/1409360.1409378}, and several models were proposed~\citep{pai-etal-2024-survey}, being either rule-based~\citep{angeli-etal-2015-leveraging} or training-based~\citep{kolluru-etal-2020-openie6}.
The extracted triples are very surface-level and noisy, with repeating, missing, or erroneous information. However, this representation enables the modeling of actions performed by entities in the text and indirectly addresses the long-range dependency problem by adopting an entity-centered perspective. In this work, two approaches to constructing the graphs were tested: \textbf{(i) OpenIE6}~\citep{kolluru-etal-2020-openie6} (not to be confused with OpenIE, the task name), an end-to-end model based on BERT and Iterative Grid Labeling for triple extraction from text. It is one of the most recent approaches with good performance and relatively low hardware needs; and \textbf{(ii)} A Mistral-Nemo model\footnote{\url{https://mistral.ai/news/mistral-nemo/}}, using the \textsc{KnowledgeGraphIndex} class of the LlamaIndex library (noted \textbf{KGI} in experiments), based on prompting an LLM in a one-shot setting.

The gist of the approach is contained in this hybrid representation, which contains both unstructured and structured representations of the data. In the following subsections, we focus on misinformation detection and propose specialized approaches that can be adapted for use in other tasks.

\subsection{TEGRA: Enriching the hybrid representation with external knowledge}
\label{sec:knowledge_incoprporation}

To enrich the constructed graph, we link the nodes in the graph with the Uniform Resource Identifiers (URIs) of the entities they contain. Tools such as DBpedia Spotlight (used in this work, based on \citep{10.1145/2506182.2506198}) and BLINK~\citep{wu2019zero} can be employed to achieve this. An external database can then be used to add information to the graph. The only requirement on this external database is that it must contain (or link to) URIs. A query language, such as SPARQL, retrieves triples from external databases and incorporates them into the constructed graph.

Initially, class-specific knowledge graphs are constructed from training documents that correspond to their labels. An OpenIE model processes the training documents. Triples extracted from documents labeled as legitimate (resp. misinformation) are stored in a graph noted $\mathcal{KG}_{true}$ (resp. $\mathcal{KG}_{misinfo}$). All triples in $\mathcal{KG}_{misinfo}$ are not necessarily false but contain possible lies and information put forward in misinformation articles. Moreover, these triples contain information pushed in fake articles’ narratives and are likely to confirm information contained in other fake articles. The Triple Selection module (noted TS) allows for modulating the influence of added information and is detailed in Subsection~\ref{sec:model_processing}.

In TEGRA, the OpenIE graph is duplicated and separately enriched by information from each KG initially constructed. For misinformation detection, we obtain two graphs $\mathcal{G}_{true}$ and $\mathcal{G}_{misinfo}$, one enriched with information from $\mathcal{KG}_{true}$ and one from information from $\mathcal{KG}_{misinfo}$. $\mathcal{G}_{true}$ only contains true information and produces more contextualized embeddings as additional true information is provided. $\mathcal{G}_{misinfo}$ does not contain only false information; however, inconsistencies can be found when comparing triples of misinformation articles with triples of legitimate articles. Additionally, the triples of misinformation articles could be aligned with other triples of misinformation, indicating that the processed text shares some common information or narratives with the training misinformation articles.

In the general case, the main idea is to enrich each graph with information from each possible class to identify similarities, correlations, or contradictions that may arise with the added information.

\subsection{Proposed hybrid models}
\label{sec:model_processing}

After constructing a hybrid graph-text representation, a specific model must be used to leverage the combined representation.

The text is processed using a standard fine-tunable encoder-based Language Model (noted LM in Figure~\ref{fig:global_archi}) such as RoBERTa~\cite{liu2019robertarobustlyoptimizedbert}.

The graph model requires several steps: \textbf{(i)} \textbf{Encoding nodes and edges.} The graph nodes are encoded using a lightweight frozen text embedding model because of the high number of nodes and memory complexity. Our implementation uses fastText~\citep{bojanowski-etal-2017-enriching} directly on nodes and edges; \textbf{(ii)} \textbf{Message passing.} Information has to propagate between nodes. To this end, we use a Graph Attention Network (GAT)~\citep{velickovic2018graph}, a lightweight and efficient model taking inspiration from the attention mechanism introduced in transformers; and  \textbf{(iii) }\textbf{Graph pooling.} After encoding each node individually, we perform maximum and mean pooling on the node embeddings and concatenate the vectors obtained (one vector per possible class). The simple graph without additional information is noted $\mathcal{G}$.

The text and involved graph embeddings are then concatenated and processed by a classification head composed of a 2-layer perceptron with two output neurons and a softmax activation function.

\paragraph{Triple Selection} Information extracted with OpenIE can be noisy or irrelevant, necessitating a mechanism to select relevant triples from the KGs, as all information related to an entity may not be relevant for the same entity in another context. To this end, we introduce a \textit{Triple Selection (TS)} module to evaluate each added triple and produce a relevance score $\mu$ between $0$ and $1$. The module takes two inputs: the fastText embeddings of the text, and the triple embeddings, which are computed as the average of the fastText encodings of the triple’s subject, predicate, and object.

Each embedding passes through its own dense projection layer before being combined with a shared dense projection layer. A dot-product and sigmoid activation function are applied to these two embeddings, producing the relevance score $\mu\in]0,1[$. The produced score is used to reduce the influence of irrelevant triples in the graph by reducing the magnitude of the embeddings of irrelevant added entities and edges before the neural graph model. In $\mathcal{G}_{true}$ and $\mathcal{G}_{misinfo}$, the corresponding added entities and edge embeddings are multiplied by the corresponding $\mu$ score. If an added entity is part of several triples, the relevance scores are averaged first. The module is not trained separately but end-to-end with the entirety of the model. 

The proposed methodology presents three families of models: \textit{text-only}, based solely on language models; \textit{TEG}, which utilizes a hybrid text-graph representation; and \textit{TEGRA}, which incorporates knowledge augmentation, which we will evaluate in the section~\ref{sec:results}.


\section{Evaluation}
\label{sec:eval}

\subsection{Datasets}

The proposed approaches were tested on four different misinformation datasets. PolitiFact and GossipCop~\citep{shu_fakenewsnet_2019}, respectively focused on fake news and rumors. CoAID~\citep{cui2020coaid} and Horne2017~\citep{Horne_Adali_2017} are topic-specific fake news datasets centered on COVID-19 and the 2016 US elections.

\subsection{Evaluation protocol}

All models are trained five times using random 80/10/10 splits for training, validation, and testing. To prevent knowledge leakage, the same data split used for the text corpus is also employed during Knowledge Base construction. Training is conducted using standard optimization practices for encoder-based models, specifically utilizing the Adam optimizer with a learning rate of $10^{-5}$. Each model is trained for up to 300 epochs, employing early stopping with a patience of 20 epochs. After training, the model with the highest F1 validation score is restored, and final performance metrics are reported on the test set. For testing metrics, we report accuracy and macro-F1 scores. All reported scores are the average of the metrics on the test set for the five splits.

\subsection{Evaluated models}
\label{sec:evaluated_models}

Three main classes of models are evaluated: Baselines, TEG, and TEGRA. Baselines are the widely used RoBERTa encoder model fine-tuned, as well as an LLM in zero- and three-shot settings. We chose Gemma3-12B, as it can be run locally on the same hardware as the proposed models, and it exhibits strong performance on various general benchmarks. In addition, we tested Tsetlin machines~\cite{bhattarai-etal-2022-explainable}, a symbolic approach based on logical rules on used words, and DeClarE (specialized in claims veracity estimation). Our proposed models are evaluated with the two graph extraction methods \textbf{OpenIE6} and \textbf{KGI}. In addition to that, TEGRA is evaluated with and without the Triple Selection module, noted \textbf{-TS} in the tables. The text backbone is roberta-base (125M parameters), and the graph models represent 724.2k additional parameters for each graph to encode (plus 270.9k for each TS).


\section{Results}
\label{sec:results}

\begin{table*}[ht]
    \centering
    \begin{tabular}{|rl||rl|rl|rl|rl|}
    \hline
        \multicolumn{2}{|c||}{\multirow{2}{*}{Configuration}} & \multicolumn{2}{c|}{PolitiFact} & \multicolumn{2}{c|}{GossipCop}& \multicolumn{2}{c|}{CoAID}& \multicolumn{2}{c|}{Horne2017}\\
        \cline{3-10}
        & & \hspace{0.4em} Acc. & F1 \hspace{1.0em} & \hspace{0.7em} Acc & F1 \hspace{1.0em} & \hspace{0.8em}Acc & F1 \hspace{1.0em}& \hspace{0.8em}Acc & F1 \hspace{1.0em}\\
         \hline\hline
         Baselines & RoBERTa & \textbf{92.97} & 91.86 & 82.68 & \textbf{83.46} & 98.42 & 95.25 & \textbf{93.33} & \textbf{94.30} \\
         & BERT & 91.00 & \textbf{91.89} & 80.25 & 78.62 & \textbf{98.72} & \textbf{95.95} & 90.91 & 91.43 \\
         & Tsetlin machine & 86.00 & 87.04 & \textbf{84.08} & 83.44 & 98.17 & 94.12 & 90.91 & 90.91 \\
         & LLM 0-shot & 78.00 & 78.85 & 66.24 & 47.52 & 84.43 & 65.86 & 84.85 & 85.71 \\
         & LLM 3-shots & 70.00 & 71.15 & 68.79 & 60.16 & 77.29 & 56.94 & 87.88 & 88.89 \\
          & DeClarE & 47.22 & 54.76 & 50.64 & 50.32 & 89.98 & 93.96 & 62.50 & 50.00 \\ 
         \hline\hline
         TEG & OpenIE6 & \textbf{95.00} & \textbf{94.68} & {77.56} & {77.56} & 99.00 & 97.06 & {92.72} & {92.49} \\
         & KGI & 93.62 & 94.08 & {76.80} & {77.16} & \textbf{99.41} & \textbf{97.75} & \textbf{96.87} & \textbf{97.88} \\
         \hline
         TEGRA & OpenIE6 & 95.00 & 94.58 & {79.23} & {79.71} & 98.85 & 96.55 & {92.12} & {92.22} \\
         & OpenIE6-TS & 95.83 & 95.52 & 82.69 & {81.87} & \textbf{99.42} & \textbf{98.30} & 93.93 & {93.75} \\
         & KGI & \textbf{96.52} & \textbf{96.39} & {82.26} & {80.21} & 99.02 & 97.10 & {90.00} & {90.81} \\
         & KGI-TS & 94.20 & 94.69 & {77.73} & {78.37} & 99.26 & 97.15 & \textbf{99.99} & \textbf{99.99}\\
         \hline
    \end{tabular}
    \caption{Misinformation detection with internal knowledge addition with different configurations. The best results for each main configuration are in bold.}
    \label{tab:results1}
\end{table*}

The results for the baselines classifiers, TEG and TEGRA for the configurations detailed in Subsection~\ref{sec:evaluated_models} are depicted in Table~\ref{tab:results1}.
Before analyzing the performance of TEG and TEGRA, it is necessary to comment on the performance achieved by generative AI for this task. Performance metrics are far from those achieved with fine-tuned encoder models, even when examples are given. Although we recognize that we only evaluated one medium-sized LLM, the performance, given the inference time, is not at the level achieved by encoders, which are trained and evaluated faster than a 12B LLM evaluated on a consumer GPU. Inference time and costs also make encoder models still a good choice for scalable systems, given their performance and fast inference for misinformation detection. DeClarE achieves good performance on CoAID (medical claims), but struggles significantly on full articles. Tsetlin machines have overall good performance, lower than that of language models, but at a significantly reduced training cost, running on CPU.

For the remainder of the paper, we consider RoBERTa as our primary baseline, given its overall mean performance across all datasets.

\subsection{Benefits of using the hybrid representation}
\label{sec:benef}

The first observation is that structuring the text in the form of a graph is not beneficial in detecting every type of misinformation. In particular, rumor detection has not been improved, indicating that stylistic features remain the most effective for detecting them.

For fake news detection, TEG leads to a mean improvement of 1.86\% F1-score over all configurations and a mean improvement of 2.97\% after selecting the best type of graph among OpenIE6 and KGI.
For PolitiFact and CoAID, TEG proved beneficial for all types of graphs. However, for Horne2017, only graphs generated with the KGI method were beneficial. This may be due to the quality and structure of the generated graphs, which differ significantly between OpenIE6 and KGI. KGI graphs are generally smaller and contain fewer components and triples; thus, fewer entities are recognized (see Appendix~\ref{app:graph_details} for details). However, the connectivity between nodes is similar for both methods. There is an exception for CoAID, as the documents are shorter, which leads to more difficulties for OpenIE6 in extracting triples. However, OpenIE6 has lighter hardware requirements, relying on a BERT backbone (108M), while KGI relies on a Mistral-Nemo LLM (12B).

\subsection{Experiments on knowledge enrichment of TEG with TEGRA}

Knowledge enrichment using information from the training dataset has been explored, demonstrating additional performance improvements in fake news detection. Incorporating this knowledge on top of previous enhancements yields an average F1-score gain of 4.42\% compared to using text alone.

Some performance differences appear depending on the choice of graph extraction method and whether the triple selection module is used or not. 

We note that there is no clear correlation between the best method for graph extraction and the use of TEG and TEGRA. However, the best-performing approach with TEGRA consistently outperforms the best approach using only TEG.

Using the triple selection module is generally beneficial and additionally improves the average F1 score by 1.51\%. Using TEGRA with triple selection for a given graph extraction method almost always improves performance over TEG, except for CoAID with KGI, which still beats the textual baseline. The primary anomaly in the results is the Politifact scores for TEGRA with KGI, which outperforms all other methods on PolitiFact. This may be due to the simple structure of the KGI graphs and the diversity of entities in the dataset, which makes the triple selection module unnecessary. Ablation studies and error analysis are provided in Appendices~\ref{sec:appendix1} and \ref{sec:appendix2}.


\section{Conclusion and Future Work}
\label{sec:conclusion}

In this paper, we present TEG, a novel approach to processing texts for knowledge-intensive tasks. We demonstrate that a hybrid text-graph representation enhances performance for fake news detection tasks and facilitates the better integration of explicit knowledge. Then, for fake news detection, we proposed TEGRA to enrich the hybrid representation with task-specific data, significantly increasing performance over standard encoder models.

However, the proposed knowledge enrichment method is task-specific and can be improved. Retrieval is currently naive and relies solely on entities. The strategies used in RAG could be leveraged to select relevant triples among texts, reducing the reliance on the TS module. Current experiments focus on incorporating knowledge external to the processed text, but closely linked to the task, as this knowledge originates from the same dataset. Future experiments will focus on adding external knowledge from a general corpus or triples from Knowledge Bases such as DBpedia~\cite{10.1007/978-3-540-76298-0_52} or YAGO~\cite{Suchanek2023YAGO4A}.

\section*{Acknowledgements}

This project was provided with computing AI and storage resources by GENCI at CINES thanks to the grant 2024-AD011015826 on the supercomputer Adastra's MI250x partition.

\section*{Limitations}

The proposed approach has been tested on a specific task using several datasets across various topics. However, the approach has not been tested in a general setting, with the same model processing all kinds of fake news on topics other than politics or COVID-19.

While strictly adhering to good practices and standards in data science, the proposed benchmark may not perfectly align with real-world use cases for misinformation detection. In production, the corpora from which the KGs are constructed should be thoroughly controlled to ensure that added information is always published before the document is processed.

\section*{Ethical statement}

Since this work addresses the topic of misinformation detection, the datasets used may contain personal attacks on public figures and occasionally include offensive content that can be harmful to their readers.

Using this type of system at scale for fake news detection should be carefully considered, and several additional experiments on the relevant data should be conducted to ensure that no potential bias (in topics, political orientation, sources, etc.) remains. We recommend evaluating the trustworthiness of the global system that would utilize our approach, using dedicated tools such as ALTAI\footnote{\url{https://digital-strategy.ec.europa.eu/en/library/assessment-list-trustworthy-artificial-intelligence-altai-self-assessment}} for the responsible use of such systems.


\bibliography{custom}

\appendix

\renewcommand\thefigure{\thesection.\arabic{figure}}
\renewcommand\thetable{\thesection.\arabic{figure}}
\setcounter{figure}{0}  
\setcounter{table}{0}  

\section{OpenIE graphs statistics}
\label{app:graph_details}

\begin{figure*}
    \centering
    \includegraphics[width=\linewidth]{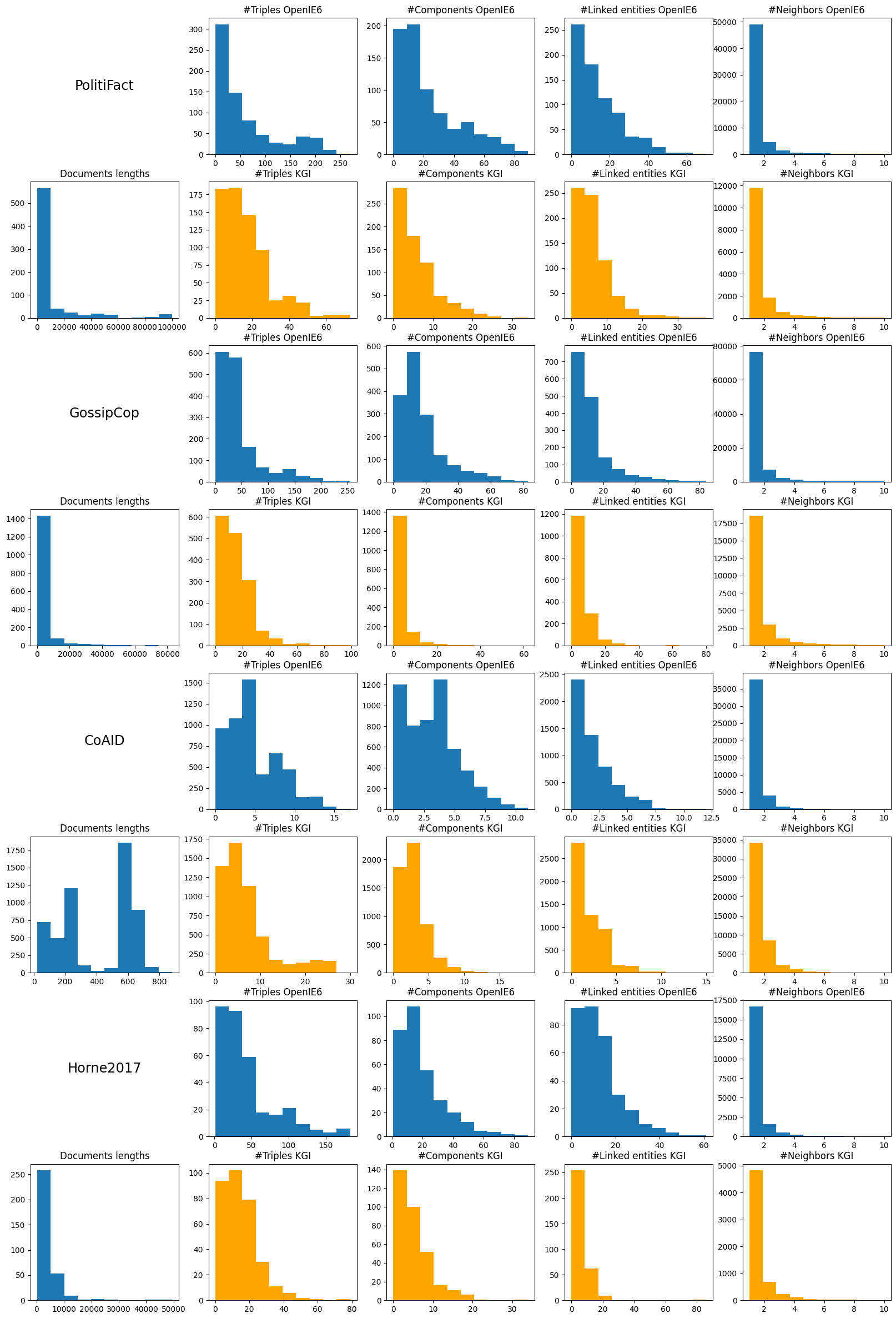}
    \caption{Datasets statistics regarding lengths, and graph sizes.}
    \label{fig:stats_graph_details}
\end{figure*}

Datasets statistics are reported visually in Figure~\ref{fig:stats_graph_details}, including for each dataset the lengths of articles in number of characters, and the number of extracted triples, graph components, linked entities and degrees of nodes (or number of neighbors) in graphs extracted by OpenIE6 and KGI.

\section{Ablation study}
\label{sec:appendix1}

The TEGRA approach comprises multiple components and utilizes several embeddings to perform classification. To better understand which embeddings are more important for classification, we performed ablation studies on the best-performing configurations for each fake news dataset (KGI for PolitiFact, OpenIE-TS for CoAID, and KGI-TS for Horne2017). Two additional experiments were conducted:

\begin{itemize}
    \item Removing $\mathcal{G}_{misinfo}$: In this configuration, only text and $\mathcal{G}_{true}$ are used, ensuring that only legitimate information is added.
    \item Removing $\mathcal{G}_{true}$: Only text and $\mathcal{G}_{misinfo}$ are used in this configuration. While this approach is well-suited to misinformation detection, it may not generalize to other tasks where knowledge about the positive class is either unavailable or challenging to define.
\end{itemize}

When removing the influence of $\mathcal{G}_{misinfo}$ and $\mathcal{G}_{true}$, we use only the text encoding, falling back to the Text-only RoBERTa baseline, which has already been evaluated.

We report the results of the ablation study in Table~\ref{tab:ablation}.

\begin{table*}[ht!]
    \centering
    \begin{tabular}{|rl||rl|rl|rl|}
    \hline
        \multicolumn{2}{|c||}{\multirow{2}{*}{Approach}} & \multicolumn{2}{c|}{PolitiFact} & \multicolumn{2}{c|}{CoAID}& \multicolumn{2}{c|}{Horne2017}\\
        \cline{3-8}
        & & \hspace{0.4em} Acc. & F1 \hspace{1.0em} &  \hspace{0.8em}Acc & F1 \hspace{1.0em}& \hspace{0.8em}Acc & F1 \hspace{1.0em}\\
         \hline\hline
         \multicolumn{2}{|c||}{TEGRA - Best configuration} & 96.52 & 96.39 & 99.42 & 98.30 & 99.99 & 99.99 \\
         \hline
         \hline
         \multicolumn{2}{|c||}{$-\mathcal{G}_{misinfo}$} & 93.04 & 93.57 & 98.75 & 96.29 & \textbf{96.25} & \textbf{97.33}\\
         \multicolumn{2}{|c||}{$-\mathcal{G}_{true}$} & \textbf{96.52} & \textbf{96.36} & \textbf{98.91} & \textbf{96.77} & 95.00 & 96.47\\
         \multicolumn{2}{|c||}{$-\mathcal{G}_{misinfo}-\mathcal{G}_{true}$ (Text-only)} & 92.97 & 91.86 & 98.42 & 95.25 & 93.33 & 94.30\\
         \hline
    \end{tabular}
    \caption{Ablation study for TEGRA on fake news datasets. It enables us to identify the types of embeddings that are most useful for the TEGRA approach. The best ablated results are in bold for each fake news dataset.}
    \label{tab:ablation}
\end{table*}

For PolitiFact and CoAID, the model relies more on $\mathcal{G}_{misinfo}$ embeddings than on $\mathcal{G}_{true}$ embeddings. The opposite is observed for Horne2017. This could be explained by the proportion of fake news in these datasets. PolitiFact and CoAID have less fake news (43\% and 17\%) in proportion than Horne2017 (61\%), making the embeddings of the minority class more discriminative.

\section{Error analysis}
\label{sec:appendix2}

To further understand the strengths and weaknesses of TEG and TEGRA, we perform an error analysis on the PolitiFact and CoAID test splits for the KGI configuration of TEG and TEGRA. These datasets and configurations were selected because they demonstrate good performance while still offering room for improvement.

On PolitiFact, no correctly classified article by RoBERTa changes the labels with TEG or TEGRA. Most incorrectly classified texts by RoBERTa are fake news misclassified as legitimate news (15 articles, with only one legitimate article incorrectly classified). TEG helps correctly classify 10 of these 15, and TEGRA adds two additional correct samples (12/15). Additionally, we observe that RoBERTa struggles more with shorter content, as most articles fill the context window; however, incorrectly classified articles have an average of 199 words. TEG works better when the extracted triples are richer and has difficulties when little information is contained in the text. Retrieval augmentation is beneficial when a large amount of false information is available.

On CoAID, there is more diversity in the labels of the wrongly classified samples. TEG corrects most of RoBERTa’s errors, but TEGRA has more difficulties, even changing one correct label from RoBERTa to a wrong one. For this corpus, RoBERTa has difficulties with longer texts. This is mitigated by the better performance of TEG when more triples can be extracted, confirming the observation made on PolitiFact. TEGRA performs better when information is scarce, and it also confirms the necessity of a triple selection module when a large amount of information is available. The change in label from RoBERTa to TEGRA is observed on a relatively longer text with only a small number of triples (only 3, which is very low for such a text of this length). This suggests that for future work, a triple quality control would be necessary to design a workflow for determining when to use TEG or TEGRA based on text and graph characteristics. 

To gain deeper insights into the strengths and weaknesses of TEG and TEGRA, we conduct an error analysis on the PolitiFact and CoAID test splits using the KGI configuration of both models. These datasets and configurations were selected because they demonstrate good performance while still offering room for improvement. The main metrics for comparison are shown in Table~\ref{tab:error_metrics}, and we detail the main observations below.

\begin{table*}[ht!]
    \centering
    \begin{tabular}{|r|c|c|}
        \hline
        Dataset & PolitiFact & CoAID \\
        \hline
        \#Words & 1216 $\pm$ 2914 & 64 $\pm$ 37 \\
        \#Words in correctly classified & 1498 $\pm$ 3237 & 63 $\pm$ 37 \\
        \#Words in uncorrectly classified & 199 $\pm$ 173 & 85 $\pm$ 36 \\
        \hline
        \multicolumn{3}{|c|}{\textbf{Correctly classified by TEG}} \\
        \hline
        \#Triples in KGI & 9.20 $\pm$7.34 & 4.50 $\pm$2.63 \\
        \hline
        \multicolumn{3}{|c|}{\textbf{Uncorrectly classified by TEG}} \\
        \hline
        \#Triples in KGI & 7.20 $\pm$3.00 & 3.0 $\pm$0.00 \\
        \hline
        \multicolumn{3}{|c|}{\textbf{Correctly classified by TEGRA}} \\
        \hline
        \#Triples in KGI & 9.17 $\pm$6.71 & 3.25 $\pm$1.09 \\
        \hline 
        \#Triples in Consistency & 50.25 $\pm$63.85 & 22.25 $\pm$28.47 \\
        \hline
        \#Triples in Contradiction & 110.25 $\pm$113.68 & 17.25 $\pm$25.27 \\
        \hline
        \multicolumn{3}{|c|}{\textbf{Uncorrectly classified by TEGRA}} \\
        \hline
        \#Triples in KGI  & 6.00 $\pm$3.27 & 4.00 $\pm$2.77 \\
        \hline
        \#Triples in Consistency & 47.33 $\pm$61.31 & 53.17 $\pm$48.97 \\
        \hline 
        \#Triples in Contradiction & 46.33 $\pm$59.90 & 46.67 $\pm$52.20 \\
        \hline
    \end{tabular}
    \caption{Graph statistics for wrongly classified examples by RoBERTa on the PolitiFact and CoAID test splits. All reported results are computed on documents that RoBERTa wrongly classifies. Displayed numbers are the mean values over the set, along with their standard deviation.}
    \label{tab:error_metrics}
\end{table*}

On the PolitiFact dataset, none of the articles correctly classified by RoBERTa had their labels changed by either TEG or TEGRA. Among the 15 articles that RoBERTa misclassified — mostly fake news labeled as legitimate, with only one legitimate article misclassified — TEG successfully corrected 10 cases. TEGRA further improved performance by correctly classifying two additional articles, bringing the total to 12 out of 15. A notable trend is that RoBERTa seems to struggle with shorter articles: while most test samples fill the context window, misclassified articles average only 199 words. TEG performs better when the extracted graphs are larger and tends to struggle when the textual information is sparse. Retrieval augmentation proves beneficial, especially in cases where the input contains a high density of false information.

On the CoAID dataset, the misclassified samples exhibit greater label diversity than those in PolitiFact. While TEG successfully corrects the majority of RoBERTa’s errors, TEGRA struggles more, in one case, even altering a correct RoBERTa prediction to an incorrect one. RoBERTa has particular difficulty with longer texts in this dataset. However, TEG mitigates this issue by performing better when more knowledge graph triples can be extracted, reinforcing the pattern observed in PolitiFact. TEGRA performs best when the additional retrieved information is limited, further supporting the need for a triple selection mechanism when large volumes of information are present. The misclassification introduced by TEGRA occurred on a relatively long article where only three triples were extracted—an unusually low number given the text length. This highlights the importance of incorporating a triple quality control mechanism in future work to guide the choice between TEG and TEGRA based on the text's characteristics and the extracted graph.

\end{document}